\documentclass[9pt,technote]{IEEEtran}

\usepackage{times}
\usepackage[T1]{fontenc}
\usepackage{amsmath}
\usepackage{bm}
\usepackage{multicol}

\usepackage{graphicx}
\graphicspath{ {images/} }

\usepackage{draftwatermark}
\SetWatermarkText{PRE-PRINT}
\SetWatermarkLightness{.85}
\SetWatermarkScale{.8}

\begin{document}

%\IEEEPARstart{W}{ith}

\title{Word embeddings for idiolect identification}

\author{\IEEEauthorblockN{
Konstantinos Perifanos\IEEEauthorrefmark{1},
Eirini Florou\IEEEauthorrefmark{2}, Dionysis Goutsos\IEEEauthorrefmark{3}}\\
\IEEEauthorblockA{ Department of Linguistics, National and Kapodistrian University of Athens
\\
\IEEEauthorrefmark{1}kostas.perifanos@gmail.com,
\IEEEauthorrefmark{2}eirini.florou@gmail.com,
\IEEEauthorrefmark{3}dgoutsos@phil.uoa.gr}}

\date{}

\IEEEoverridecommandlockouts
\IEEEpubid{\makebox[\columnwidth]{978-1-5386-8161-9/18/$31.00~\copyright2018 IEEE \hfill} \hspace{\columnsep}\makebox[\columnwidth]{ }}

\maketitle

%\IEEEpubidadjcol

\begin{abstract}
The term idiolect refers to the unique and distinctive use of language of an individual and it is the theoretical foundation of Authorship Attribution. In this paper we are focusing on learning distributed representations (embeddings) of social media users that reflect their writing style. These representations can be considered as stylistic fingerprints of the authors. We are exploring the performance of the two main flavours of distributed representations, namely embeddings produced by Neural Probabilistic Language models (such as word2vec) and matrix factorization (such as GloVe).
\end{abstract}

\thispagestyle{empty}

\section{Introduction}

The introduction of the term {\itshape idiolect} in sociolingustics appears in the 19th century by Neogramarian Herman Paul \cite{paul1890principles}. The theoretical concept and the existence of idiolect is the foundation of Authorship Attribution \cite{coulthard2004author}. Here, we introduce a new way of capturing and quantifying author style, by producing an author's "stylistic fingerprint" using word embeddings. Our method is not limited by the corpus vocabulary size and it is scalable to thousands of authors.

\section{Motivation and Related Work}

Stylistic similarity has various practical applications, including authorship attribution (given a disputed text between two or more authors, identify the author of the text) \cite{coyotl2006authorship}, plagiarism detection \cite{maurer2006plagiarism}, online harassment and abuse \cite{tan2013unik}. Applications of stylistic similarity are not limited in natural language problems and text; they are also applied in similar problems related to source code \cite{burrows2007source}, music scores \cite{van2005musical} etc. Similar techniques can also be considered as extensions to content recommendation systems, for example for news sites, literature recommendations as well as sub-components of content personalisation engines.

There are several approaches in the literature trying to address the problem of stylistic similarity, from a computational-corpus linguistic point of vie1w.

Barlow \cite{barlow2013individual} examines the corpus of five White House Secretaries and focuses the analysis on the distribution of bigrams and trigrams per author.

Mollin \cite{mollin2009entirely} focuses on the individual style of Tony Blair in comparison with British National Corpus (B.N.C.). She compares the Tony Blair Corpus (T.B.C.), which consists of Tony  Blair's speeches and interviews and quantifies the difference between the usage frequency of maximizer collocations such as "I entirely understand".

Hughes et.al \cite{hughes2012quantitative} are investigating writing style in Literature, by examining 537 authors from Project Gutenberg. They analyze 7337 works, using the frequency of 307 content-free words. They generate feature vectors for each author based on the frequency of the content-free words and they are using symmetric Kullback–Leibler divergence as a distance measure.

In our approach, we are expanding and removing the limitations on the studies above, both in size of the vocabulary as well as in the number of authors. Our method is applied on large vocabulary sizes and not only top N stop words.
Our method is applied and proven to be consistent in two corpora:
A selection of tweets of 4494 authors, consisting of approximately 302 million words and corpus of blog posts consisting of 19320 authors and 140 million words.

\section{Corpora}

Our work on idiolect and writing style focuses on Social Media, and more specifically on Twitter and blogs. Sociolinguistics research on Twitter and related Social Media Platforms is quite common in the literature for various reasons: volume of data available as well as variability in gender, geolocation etc. \cite{bamman2014gender,mikros2013authorship,schwartz2013personality}

For the purposes of the work presented in this paper, we collected timelines from 4494 twitter users, mostly tweeting in Greek language.
The corpus consists of 26103963 tweets, tweeted between 2008 and 2017 and 325,243,302 words (whitespace tokenized).

We also applied our methodology in the The Blog Authorship Corpus by Schler et. al. \cite{schler2006effects}.

\section{Word and Paragraph Embeddings}

Word embeddings are essentially representations of words in a d-dimensional vector space. In the most typical scenario, embeddings are learned by applying Neural Probabilistic Language Models. Neural Language Models were introduced by Bengio et. al. \cite{bengio2003neural} in 2003. They became popular and mainstream after the publishing of word2vec, an implementation of neural language model by Mikolov et. al.\cite{mikolov2013efficient}. The word2vec tool provided the ability for fast training in large corpora (hundreds of millions of words). The produced embeddings have a very nice feature: Semantically related words tend to be close in the d-dimentional space. wor2vec introduces two architectures, Continuous Bag of Words (CBOW), where the neural network tries to predict the next word given context, and Skip-Gram where given the input word the model tries to predict the context. To speed up calculation, Mikolov et. al. \cite{NIPS2013_5021} introduce Negative Sampling, a simplified variant of Noise-contrastive estimation \cite{gutmann2010noise}. In this scheme, instead of trying to calculate the softmax of all words in the vocabulary, a fixed number $K << |V|$ of negative samples are drawn from the vocabulary.

Following word2vec Le and Mikolov introduce the distributed representation of sentences and documents (Paragraph Vectors) \cite{le2014distributed}. This is a natural extention of word2vec, where metadata on sentence or even document level are treated as additional words/tokens and then are fed  as additional input to the neural network. The resulting outcome is not only d-dimentional vectors for vocabulary words but also d-dimentional vectors for the sentence level metadata. Paragraph Vectors comes also in two flavours: Distributed Memory Model of Paragraph Vectors (PV-DM) and Distributed Bag of Words version of Paragraph Vector (PV-DBOW). PV-DM preserves word order in the context where PV-DBOW ignores word ordering.

\section{Experiment and results}

In our work, we are learning Paragraph Vectors on author level. More concretely, we aggregate all tweets by the same user in a single document and we are feeding words and author id's to the neural model. The corpus is preprocessed before training. Words are whitespace tokenized, all tweets are lowercased and we are removing @ mentions. The reason mentions are removed is because mentions are carrying information on social network interactions among users and they can be used to model user proximity in a social graph \cite{schaal2012analysis}. Since we want to learn user similariy based on vocabulary only choices, we are discarding all social interactions and we are focusing on language usage only.

Because we are interested in stylistic choices, word order is important and thus we are focusing only on the PV-DM model.	As described by Mikolov et. al, the author token acts as memory that remembers what is missing from the current context. \footnote{The implementation is based on the Gensim Python library \cite{rehurek2011gensim}, https://radimrehurek.com/gensim/}

For evaluation purposes, we created a random subset of initially 100 users from the corpus. For each user $user_i, i \in \{1,2,...100\}$ in this set, we randomly splitting the tweets in two sets, $user_i\_A$, $user_i\_B$, each set containing 50\%  of the user's tweets.

We then train the model and we are evaluating in the following way: We are extracting the distributed representation all users $user_i$ and we querying for most similar users against all learned distributed author representations. If  $user_i\_A$ is the most similar to $user_i\_B$ this is considered as a positive match, negative otherwise. We are repeating this experiment for various values of embedding dimentionallity $D$ and calculate the accuracy as described above.

\begin{figure}
  \includegraphics[width=\linewidth]{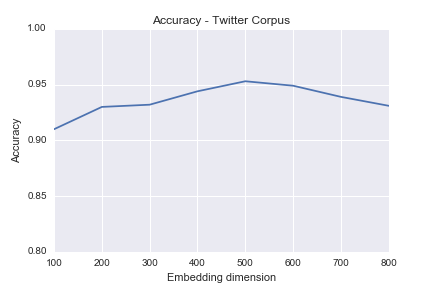}
\end{figure}

The accuracy drops in cases where a user has less than 500 tweets, which is a reasonable outcome given the fact that due to 140 character limitation in twitter statuses until 2017, the total words for 500 tweets are not enough to discriminate user style.

We are then repeating the process above by gradually increasing the number of twitter users to 4000. Accuracy remains consistent and always higher than 85\%. The drop in accuracy is explained by the fact that the words/tweets distribution follows Zipfian distribution and as we are increasing the sample size for the test set, the probability of selecting a user with less than 500 tweets is increasing.

If we relax the definition of a positive match to "user in top K most similar tweets", we achieve accuracy always near to 100\%. This is of huge practical importance in cases of Forensic Linguistics such as Doppelg{\"a}nger  Detection \cite{afroz2014doppelganger}, where learned representations can be used as features in relevant models along with social graph information as well as metadata (Device, user habits etc).

\section{Sociolect, style stability, applications and future work}

Subsequently, given the vector representation of authors, we can apply clustering techniques to formulate clusters of "sociolect" and group users of similar or near identical writing style. This also reflects popularity, especially on twitter, where users tend to adopt trending words, phrases or slang words introduced by other popular users, in different contexts: From casual chat to marketing, politics and propaganda \cite{goodenow1989hiv}. This method then can be used as a "who to follow" recommendation strategy for users with similar styles. One other potential direction is to apply network backbone analysis \cite{serrano2009extracting} to identify the most stylistically influential authors in the coprus.

One important question we are also invastigating is an individual's idiolect stability over time. Again, by applying the same methodology as above, we are splitting user tweets in pairs (user,year) and we are querying and comparing most similar users to a given user\_A  in a given year, in all previous years. The results are promising, as we are achieving again high accuracy scores, near 80\%. However, since now we are splitting users in multiple (user, year) pairs, accuracy drops. One obvious way to address that is to collect more data and repeat the experiments with more authors and tweets.

Finally, we aim to apply the same methodology on other text or text-like collections, such as News and Literature.

\bibliographystyle{IEEEtran}
\nocite{*}

\bibliography{idiolect_embeddings}

\end{document}